\documentclass[letterpaper, 10 pt, conference]{ieeeconf}  

\IEEEoverridecommandlockouts                              

\usepackage[colorlinks,linkcolor=blue]{hyperref} 
\usepackage{mathrsfs}
\usepackage{amsfonts}
\usepackage{amsmath}
\usepackage[flushleft]{threeparttable}
\usepackage{tablefootnote}
\usepackage{multirow}
\usepackage{graphicx}
\usepackage{subfigure}
\usepackage{hanging}
\usepackage{color}
\usepackage{tikz}
\usetikzlibrary{calc,arrows,decorations.markings}
\usepackage{balance}
\usepackage{cite}
\usepackage{booktabs}
\usepackage{xpatch}
\usepackage{mathrsfs}
\usepackage{amsmath}
\DeclareMathOperator*{\argmax}{arg\,max}

\makeatletter
\patchcmd\@makecaption{\\}{.~}{}{\fail}
\makeatletter

\title{\LARGE \bf
Task-Oriented Grasp Prediction with Visual-Language Inputs}

\author{Chao Tang$^{1}$, Dehao Huang$^{1}$, Lingxiao Meng$^{1}$, Weiyu Liu$^{2}$, Hong Zhang$^{1}$ \emph{Fellow, IEEE}
\thanks{$^{1}$Shenzhen Key Laboratory of Robotics and Computer Vision, Southern University of Science and Technology, Shenzhen, China.}%
\thanks{$^{2}$Institute for Robotics and Intelligent Machines, Georgia Institute of Technology, Atlanta, United States.}%
}

\begin{document}

\maketitle


\begin{abstract}

To perform household tasks, assistive robots receive commands in the form of user language instructions for tool manipulation. The initial stage involves selecting the intended tool (i.e., object grounding) and grasping it in a task-oriented manner (i.e., task grounding). Nevertheless, prior researches on visual-language grasping (VLG) focus on object grounding, while disregarding the fine-grained impact of tasks on object grasping. Task-incompatible grasping of a tool will inevitably limit the success of subsequent manipulation steps. Motivated by this problem, this paper proposes GraspCLIP, which addresses the challenge of task grounding in addition to object grounding to enable task-oriented grasp prediction with visual-language inputs. Evaluation on a custom dataset demonstrates that GraspCLIP achieves superior performance over established baselines with object grounding only. The effectiveness of the proposed method is further validated on an assistive robotic arm platform for grasping previously unseen kitchen tools given the task specification. Our presentation video is available at: \href{https://www.youtube.com/watch?v=e1wfYQPeAXU}{https://www.youtube.com/watch?v=e1wfYQPeAXU}.

\end{abstract}

 
\thispagestyle{empty}
\pagestyle{empty}

\section{Introduction}

Language provides a natural interface for task specification in unconstructed environments such as kitchens and offices, complementing pure vision-based robotic frameworks \cite{mahler2017dex, mousavian20196, chu2018real}. Guided by natural language, an assistive robot is able to perform a wide range of household manipulation tasks using verbal instructions, such as ``Use the \textit{knife} to \textit{cut} the apple for me" and ``\textit{Clean} the mug with a \textit{brush}". The initial step in performing such tasks is to grasp the intended tool in a task-oriented manner. This necessitates that the robot both coarsely localizes the target object (i.e., object grounding) and comprehends which fine-grained object part to grasp for the intended task execution (i.e., task grounding). However, previous researches on VLG, such as natural language object retrieval\cite{hatori2018interactively, shridhar2018interactive,zhang2021invigorate} and object rearrangement \cite{liu2022structformer}, focus on grounding language instructions to some coarse object-centric representations (e.g., bounding box, instance segmentation mask), while disregarding the fine-grained, task-oriented effects on object grasping. 

\begin{figure}[t]
  \centering
  \begin{tikzpicture}[inner sep = 0pt, outer sep = 0pt]
    \node[anchor=south west] (fnC) at (0in,0in)
      {\includegraphics[height=3.6in,clip=true,trim=0in 0in 0in 0in]{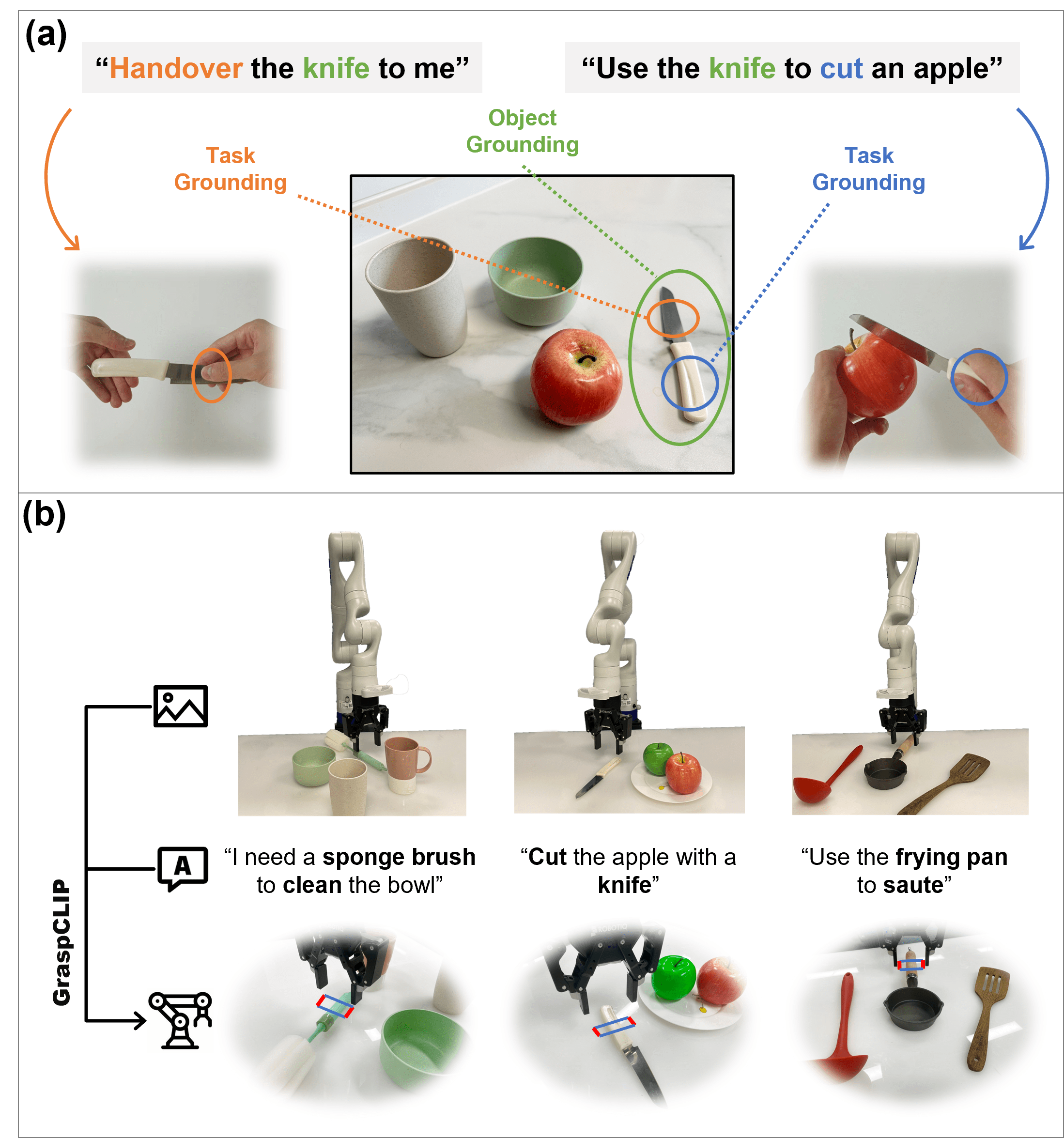}};
  \end{tikzpicture}
    \vspace*{-0.25in}
  \caption{(a) Task grounding and object grounding revealed in humans' grasping behavior. (b) GraspCLIP takes as input a visual scene observation $O$ of multiple objects and a task instruction $I$, and outputs a task-oriented grasp pose $g$.}
  \label{fig:concept}
  \vspace*{-0.3in}
\end{figure}

Fig.\ref{fig:concept}(a) illustrates an example of manipulating kitchen tools. The language instruction of ``Use the \textit{knife} to \textit{cut} an apple" necessitates both grounding the target object ``\textit{knife}" and grounding the target task of ``\textit{cut}" to the handle of the knife. Conversely, when the language instruction is ``\textit{Handover} the \textit{knife} to me", humans would choose a different way by holding the blade for the target task of ``\textit{handover}". It is clear from this example that a language instruction would affect not only what object to grasp but also how the target object is grasped for an intended task execution. We, humans, take this skill for granted, but it is not explored by previous VLG researches. Disregarding the fine-grained effects of tasks on grasp poses may result in potential task failures. For instance, handover by grabbing the knife handle may cause physical injury to the receiver. Furthermore, imprecise grasping of the knife handle may result in cutting failure. So, how can we endow robots with the same ability to predict task-oriented grasps with visual-language inputs?

To answer this question, we propose GraspCLIP to address task grounding in addition to object grounding to enable task-oriented grasp prediction. Fig.\ref{fig:concept}(b) shows an assistive robot operating in a kitchen environment. GraspCLIP takes as input a visual scene observation $O$ of multiple objects and a task instruction $I$, and outputs a task-oriented grasp pose $g$. GraspCLIP first leverages a visual-language model (VLM) CLIP\cite{radford2021learning} pre-trained on large-scale internet data to encode multi-modal inputs into a joint representation space. Then, to simultaneously achieve task grounding and object grounding, a two-stage, coarse-to-fine Task-Oriented Fusion (TOF) module is proposed to build hierarchical correspondences between visual observations and task instructions. This is in contrast to previous works, which have focused only on object grounding. In the last stage, a decoder predicts task-oriented grasp poses based on instruction-conditioned representations generated from the previous stage. Evaluation on a custom dataset demonstrates the superiority of GraspCLIP over established baselines with object groudning only. We further validate its effectiveness in real-world applications on an assistive robotic arm platform for grasping previously unseen kitchen tools given the task specification.

In summary, our contributions are as follows:
\begin{itemize}
    \item  To address the challenge of task grounding in addition to object grounding, we contribute GraspCLIP to enable task-oriented grasp prediction with visual-language inputs.
    \item To evaluate the task-oriented grasp prediction performance, we provide a custom dataset comprising 28 object categories, 96 instances, 38 household tasks, task-oriented grasp annotations, and template-based language instructions.
    \item A system is built to enable an assistive robotic arm to predict and execute task-oriented grasps guided by user language instructions.
\end{itemize}


\section{Related Work} \label{related_works}


Vision-based grasping (VG) has been a fundamental problem in robotics. With the rise of deep learning in recent years, VG has achieved significant advances. For example, Mahler et al. \cite{mahler2017dex} and Chu et al. \cite{chu2018real}  use CNN-based networks to predict planar grasps from RGB-D images. Mousavian et al. \cite{mousavian20196} propose to generate 6 degree-of-freedom (DoF) grasp poses on point clouds with a variational autoencoder. Most works in VG consider task-agnostic grasping, which finds stable grasp poses satisfying form and force closure. Failure to consider task constraints limits their usage in many application scenarios. 

To address this problem, some recent researches have proposed to merge language grounding into vision-based manipulation and grasping pipelines \cite{shridhar2020alfred, liu2021structformer, ahn2022can, shridhar2022cliport, chen2021joint, hatori2018interactively, shridhar2018interactive, zhang2021invigorate}. Conditioned on language, the robot can understand and execute a diverse range of VLG tasks. Hatori et al. \cite{hatori2018interactively} present the first system to resolve ambiguity in language instructions for object picking. Similarly, Shridhar et al. \cite{shridhar2018interactive} interactively pick objects using referring expressions. Built on top of \cite{hatori2018interactively} and \cite{shridhar2018interactive}, Zhang et al. \cite{zhang2021invigorate} address language-conditioned picking in the clutter. The above methods focus on grounding natural language to coarse object-centric representations such as bounding boxes and use off-the-shelf task-agnostic grasp detectors. They do not explicitly consider the fine-grained effects of tasks on object grasping. This effect is essential since a task instruction would affect not only what object to grasp but also how to grasp it for the subsequent task execution. Another problem is that they rely on deep learning models trained on small-scale, self-collected datasets or public datasets such as RefCOCO \cite{kazemzadeh2014referitgame}, limiting their generalization capability to novel scenes, instances, categories, and tasks.

There has been a recent trend of building VLG pipelines based on large pre-trained models to improve the generalization capability. For example, SayCan\cite{ahn2022can} and CLIPort\cite{shridhar2022cliport} explore the power of large pre-trained models from natural language processing (NLP) and computer vision (CV) communities to build priors for robots efficiently. Ahn et al. \cite{ahn2022can} combine low-level skills with large language models (LLMs) \cite{brown2020language}\cite{chowdhery2022palm} to complete long-horizon, language-guided mobile manipulation tasks. Shridhar et al. \cite{shridhar2022cliport} present a CLIP \cite{radford2021learning} based imitation-learning agent trained for solving various language-specified tabletop tasks. Despite demonstrating a capacity to solve complex VLG tasks, they do not explicitly consider the fine-grained effects of tasks on object grasping. As a supplement to
previous VLG researches, we address the challenge of task grounding in addition to object grounding to enable task-oriented grasp prediction with visual-language inputs.

\begin{figure}[t]
  \centering
  \vspace*{-0.1in}
  \begin{tikzpicture}[inner sep = 0pt, outer sep = 0pt]
    \node[anchor=south west] (fnC) at (0in,0in)
      {\includegraphics[height=1.2in,clip=true,trim=0in 0in 0in 0in]{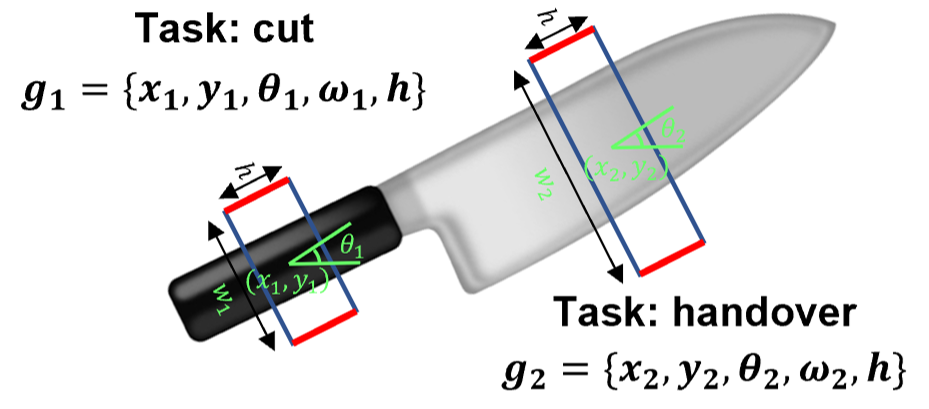}};
  \end{tikzpicture}
  \caption{Two examples of 5D grasp rectangles. The representation describes the grasp location, orientation, opening width, and length of a parallel jaw gripper. Each example is additionally annotated with task labels.}
  \label{fig:grasp_example}
  \vspace*{-0.25in}
\end{figure}

\section{Problem Formulation} \label{formulation}
We consider the problem of learning a function $\mathcal{F}$  that receives a visual scene observation $O \in \mathbb{R}^{H\times W\times3}$ and a task instruction $I=\{{s_t}\}_{t=1}^{T}$, and outputs a task-oriented grasp pose $g$ (in the image space), where $s_t$ is the $t$-th word token and $T$ is the max length:
\begin{equation*}
    g = \mathcal{F}(O, I)
\end{equation*}
Here, $O$ is an RGB image of multiple objects, including one or more target objects and distractors. $I$ is a natural language sentence of the task description. Depicted in Fig.\ref{fig:grasp_example} are two examples of $g$. Each of them is a 5-dimensional grasp rectangle parameterized by grasp location $(x, y)$, orientation $\theta$, opening width $w$, and length $h$:
\begin{equation*}\label{eq:2}
    g = \{x, y, \theta, w, h\}
\end{equation*}
where the first three parameters are in $SE(2)$ and represent the reference frame of the rectangle, and the last two describe the dimensions. $h$ is a fixed value for a designated gripper in our implementation, although it could be a learnable parameter in general. For orientation, the space of $SO(2)$ rotation is discretized into 120 bins.

\begin{figure*}[th]
  \centering
  \vspace*{-0.2in}
  \begin{tikzpicture}[inner sep = 0pt, outer sep = 0pt]
    \node[anchor=south west] (fnC) at (0in,0in)
      {\includegraphics[height=4.0in,clip=true,trim=0in 0in 0in 0.0in]{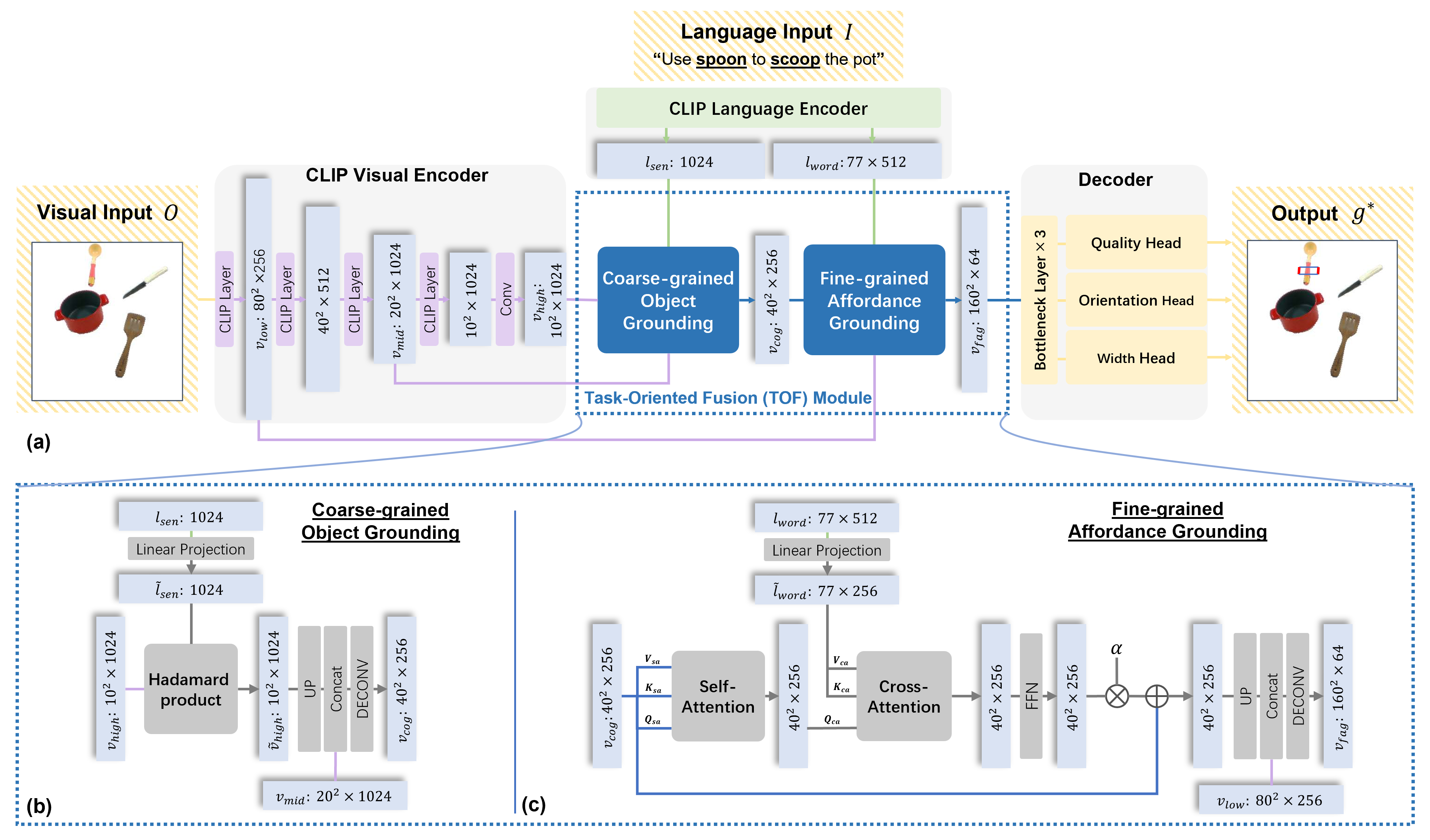}};
  \end{tikzpicture}
    \vspace*{-0.1in}
  \caption{An overview of GraspCLIP architecture: (a) GraspCLIP consists two CLIP-based encoders, a Task-Oriented Fusion module, and a decoder. (b) Coarse-grained Object Grounding module coarsely localizes the target object. (c) Fine-grained Affordance Grounding module creates fine-grained correspondences between functional/affordance regions and task instructions.}
  \label{fig:pipeline}
  \vspace*{-0.20in}
\end{figure*}

We approximate function $\mathcal{F}$ with a deep neural network, namely GraspCLIP. To train GraspCLIP, a dataset $\mathcal{D}=\{d_1, d_2,..., d_n\}$ of $n$ tuples is required. The detail of data generation will be introduced later. Each tuple consists of a visual scene observation $O_j$, a task instruction $I_j$, and a set of $m$ task-oriented grasp annotations $\mathcal{G}_j=\{g_{i, j}\}_{i=1}^{m}$:
\begin{equation*}
    d_j = (O_j, I_j, \mathcal{G}_j)
\end{equation*}
where $j=1,2,...,n$.

\section{Approach} \label{approach}

An overview of the proposed GraspCLIP is presented in Fig.\ref{fig:pipeline}. The model architecture consists of four major components: two CLIP-based encoders, a two-stage, coarse-to-fine TOF module, and a decoder. Each component will be introduced for the rest of this section.
 
\subsection{Encoder Module}
\begin{figure}[t]
  \centering
  \begin{tikzpicture}[inner sep = 0pt, outer sep = 0pt]
    \node[anchor=south west] (fnC) at (0in,0in)
      {\includegraphics[height=1.5in,clip=true,trim=0.0in 0.2in 0.1in 0.2in]{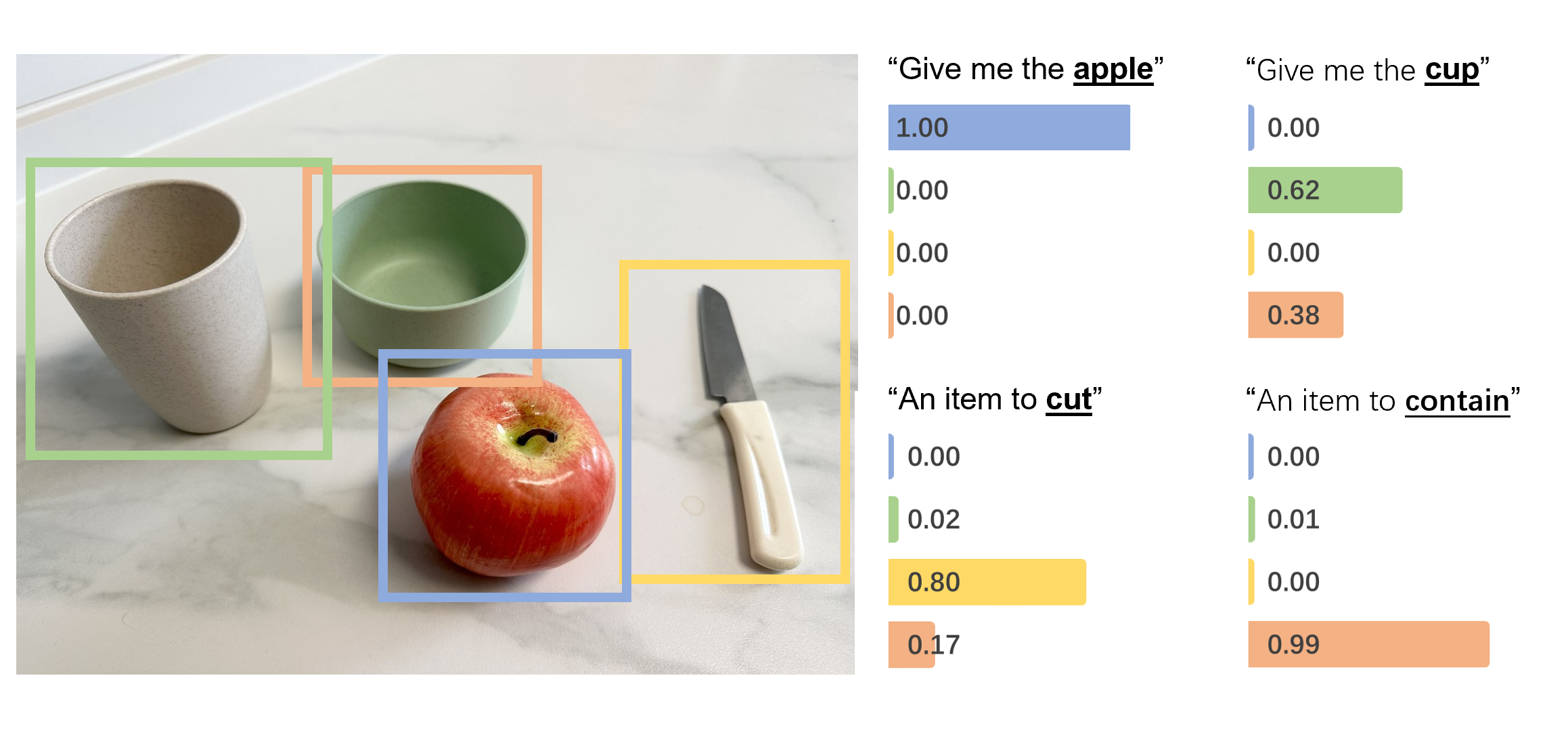}};
  \end{tikzpicture}
    \vspace*{-0.28in}
  \caption{Given detected object proposals and natural language descriptions, CLIP outputs distributions over proposals without training.}
  \label{fig:clip}
  \vspace*{-0.25in}
\end{figure}

Given a visual scene observation $O$ and a task instruction $I$, GraspCLIP first encodes multi-modal inputs into a joint representation space, which enables cross-modal reasoning and semantic understanding. Previous works on VLG usually use backbone networks trained on small-scale, single-modal datasets. This will lead to (1) a limited generalization capability to novel concepts and (2) a large semantic gap between two modalities. We, therefore, opt for CLIP-based encoders. Two encoders are pre-trained jointly on a dataset of 400 million (image, text) pairs and inherently learn a broad context of semantics. They contain rich prior knowledge for grounding open-end, high-level semantic concepts (see Fig.\ref{fig:clip}). This property is beneficial since an assistive robot in real-world applications needs to deal with a open set of object categories and tasks. Specifically, we use a CLIP pre-trained ResNet50 \cite{he2016deep} and a CLIP pre-trained BERT\cite{devlin2018bert} to encode $O$ and $I$, respectively.

Although CLIP-based encoders provide a strong basis, CLIP is originally designed to align the whole image (instead of pixels or regions) with the input sentence, leading to a significant gap between high-level image understanding and low-level task-oriented grasping. We next address this problem with a multi-modal fusion module and a decoder to transfer CLIP encodings to a task-oriented grasp prediction.

\subsection{Task-Oriented Fusion Module}\label{VLRF}
Using a single output from each CLIP encoder is not enough for accurate task-oriented grasp prediction since a task instruction contains information at multiple levels of granularity. For example, ``Use the \textit{knife} to \textit{cut} an apple" requires both coarsely grounding the target object ``\textit{knife}" and understanding which fine-grained object part to grasp for the target task of ``\textit{cut}". To tackle this issue, a hierarchical approach is first employed to capture the semantic meaning of multi-modal inputs. First, we transform $I$ into two types of language embeddings: a sentence embedding vector $l_{sen}  \in \mathbb{R}^{1024}$ and a word embedding sequence $l_{word} \in \mathbb{R}^{77\times 512}$ (with zero-padding). While $l_{sen}$ provides a broad abstraction of the whole instruction, $l_{word}$ stores an detailed embedding vector for each individual word token. Similarly, the intermediate features from CLIP visual encoder are also extracted to obtain a hierarchical representation of $O$ (i.e., object-part-shape). To achieve both object grounding and task grounding, we then need to build hierarchical correspondences between two sets of representations. Thus, a two-stage, coarse-to-fine Task-Oriented Fusion (TOF) module is proposed. It consists of a Coarse-grained Object Grounding (COG) module and a Fine-grained Affordance Grounding (FAG) module. 

In the first stage, COG creates a coarse mapping from $I$ to the target object in $O$. It takes as input the high-level visual feature map $v_{high} \in \mathbb{R}^{10 \times 10\times 1024}$ and the sentence embedding $l_{sen}$. To reduce the semantic gap between two modalities, a linear projection is first applied: $l_{sen}\rightarrow \Tilde{l}_{sen} \in \mathbb{R}^{1024}$. Hadamard product is then taken at each spatial location to perform object grounding:
\begin{equation*}
    \Tilde{v}_{high, i} = v_{high, i} \odot \Tilde{l}_{sen}, i=1,...,10\times10
\end{equation*}
$\Tilde{v}_{high}$ is upsampled and concatenated with mid-level visual feature map $v_{mid} \in \mathbb{R}^{20 \times 20 \times 1024}$, followed by a transposed convolution block to output $v_{cog} \in \mathbb{R}^{40 \times 40 \times 256}$. 


However, as discussed before, object grounding is nevertheless insufficient to predict task-oriented grasps. The model must also establish a fine-grained correspondence between the target task in $I$ and a functional/affordance region on the target object. To tackle this, FAG module is introduced in the second stage. According to the theory of affordance\cite{gibson1977theory}, affordance is defined in the second order here. For example, affordances ``\textit{cut}" and ``\textit{handover}" correspond to the knife handle and blade, respectively. The architecture of FAG is shown in Fig.\ref{fig:pipeline}(c). The computational procedure can be divided into two steps. FAG first explores affordance regions on $v_{cog}$ and then maps $l_{word}$, especially object and task tokens, to these regions. To model the fine-grained intra-modal (visual affordance exploration) and inter-modal (word-to-affordance mapping) interactions, two types of Transformer-based attention mechanisms\cite{vaswani2017attention} are utilized in cascade. 

\begin{figure}[t]
  \centering
  \vspace*{-0.2in}
  \begin{tikzpicture}[inner sep = 0pt, outer sep = 0pt]
    \node[anchor=south west] (fnC) at (0in,0in)
      {\includegraphics[height=2.2in,clip=true,trim=0.2in 0.0in 0.2in 0.3in]{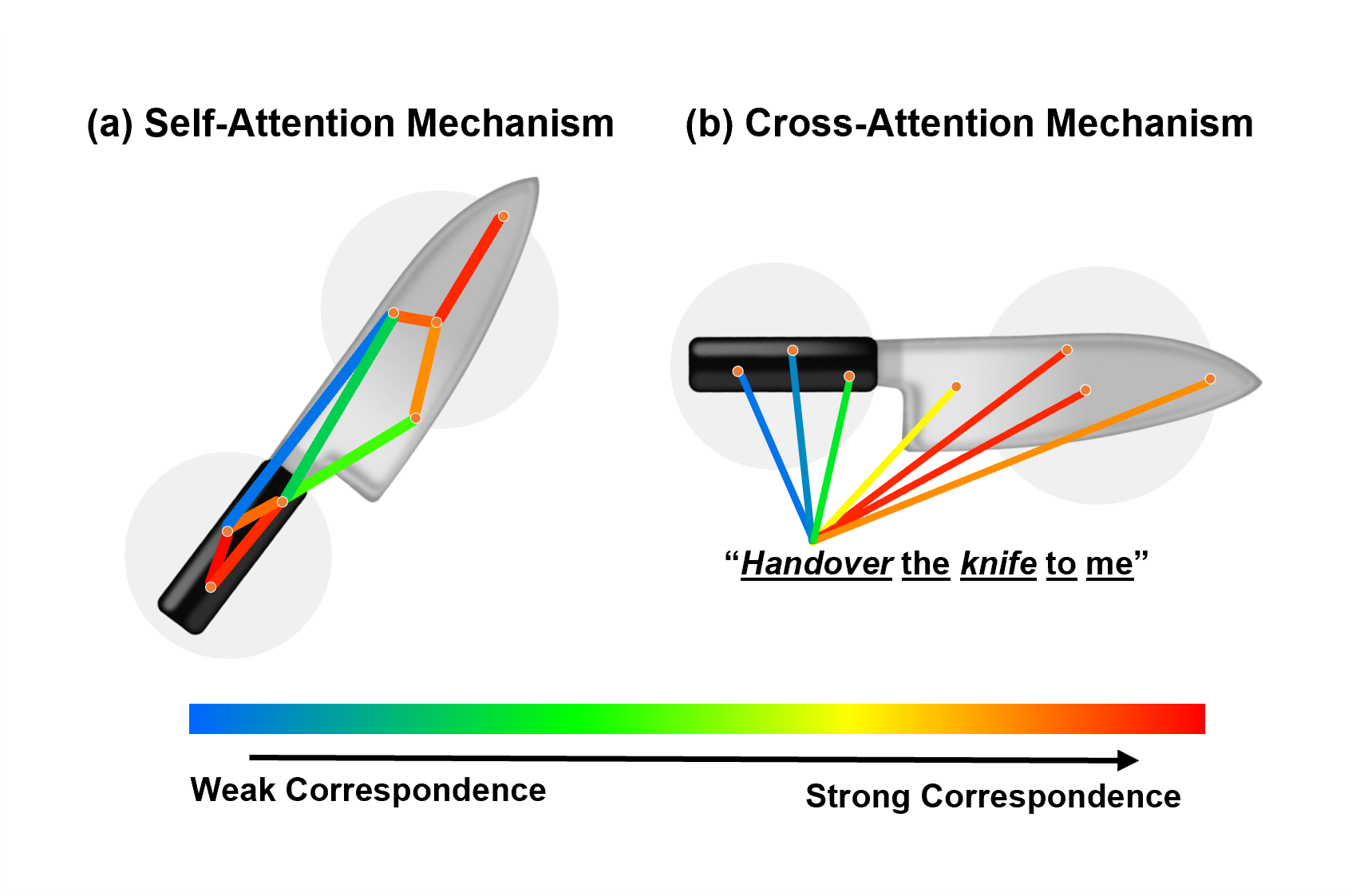}};
  \end{tikzpicture}
    \vspace*{-0.2in}
  \caption{Visualizations of two attention mechanisms.}
  \label{fig:attention}
  \vspace*{-0.2in}
\end{figure}

According to \cite{myers2015affordance}, regions sharing similar geometric structures are likely to have the same affordance. Therefore, we incorporate a self-attention layer to capture the non-local structural information on $v_{cog}$. An example is depicted in Fig.\ref{fig:attention}(a) for clarification. Self-attention provides a global context for local point-wise affordance exploration and parses $v_{cog}$ into a set of functional regions. Specifically, $v_{cog}$ is first flattened into $z_{cog} \in \mathbb{R}^{1600 \times 256}$. The self-attended feature map $z_{sa}$ can be then computed as:
\begin{gather*}
    z_{sa} = \text{softmax}(\frac{Q_{sa}K_{sa}^{\top}}{\sqrt{256}})V_{sa}, \\
     Q_{sa} = W_{sa}^{Q}z_{cog},   K_{sa} = W_{sa}^{K}z_{cog},   V_{sa} = W_{sa}^{V}z_{cog}
\end{gather*}
where $W_{sa}^{Q}$, $W_{sa}^{K}$, and $ W_{sa}^{V}$ are self-attention query, key, and value projection matrices, respectively. After the visual affordance exploration, we are ready to build fine-grained correspondences between affordance regions and word tokens. A cross-attention layer is adopted. As is shown in Fig.\ref{fig:attention}(b), the intuition is to reconstruct $z_{sa, i}$ by all elements in $l_{word}$ weighted by their normalized cross-modal correspondences, where $i=1,...,1600$. To match the dimension of $z_{sa}$, $l_{word}$ is first projected to a lower dimension of 256: $l_{word}\rightarrow \Tilde{l}_{word} \in \mathbb{R}^{77 \times 256}$. The cross-attended feature map $z_{ca}$ then can be computed as:
\begin{gather*}
     z_{ca} = \text{softmax}(\frac{Q_{ca}K_{ca}^{\top}}{\sqrt{256}})V_{ca}, z_{ca} = \text{FFN}(z_{ca}),\\
     Q_{ca} = W_{ca}^{Q}z_{sa},   K_{ca} = W_{ca}^{K}\Tilde{l}_{word},   V_{ca} = W_{ca}^{V}\Tilde{l}_{word}
\end{gather*}
where $W_{ca}^{Q}$, $W_{ca}^{K}$, and $ W_{ca}^{V}$ are cross-attention query, key, and value projection matrices, respectively. FFN is a feedforward layer. Since the training of two attention layers is computationally unstable at the beginning, we insert them with a learnable gating parameter $\alpha$ initialized to 0. In this way, GraspCLIP learns to localize the target object in the initial training stage, and gradually attends to fine-grained affordance regions supporting the target task. Finally, $z_{ca}$ is reshaped to $v_{ca} \in \mathbb{R}^{40\times40\times256}$, and then fused with low-level visual feature map $v_{low}\in \mathbb{R}^{80\times80\times256}$ to output $v_{fag} \in \mathbb{R}^{160 \times 160 \times 64}$.

\subsection{Decoder Module}
The decoder predicts a task-oriented grasp pose based on $v_{fag}$. Specifically, three consecutive bottleneck layers are first applied to output $v_{pred} \in \mathbb{R}^{640\times640\times16}$. The grasp prediction is then divided into three parallel tasks, and each is solved by appending a prediction head to $v_{pred}$. For quality head, it outputs a heatmap $M_{q} \in \mathbb{R}^{H\times W}$, measuring the probability (between 0 and 1) of satisfying the task instruction at each spatial location $(x, y)$. The other two heads output the orientation map $M_{\theta} \in \mathbb{R}^{ H\times W\times l}$ and opening width map $M_{w} \in \mathbb{R}^{H\times W}$, respectively. 

A task instruction may correspond to multiple ground truth grasp poses. Here, GraspCLIP only outputs the top-1 prediction during inference by first taking the argmax over the smoothed $M_{q}$ and then querying the other two maps:
\begin{align*}
    x^{*},y^{*} = \argmax_{x, y} \ & \text{Gaussian}(M_{q}) \\
    \theta^{*} = \argmax_{dim=2}M_{\theta} |_{(x^{*},y^{*})} , \ \  &w^{*} = \text{Gaussian}(M_{w})|_{(x^{*}, y^{*})} 
\end{align*}
The output grasp pose $g^{*}$ is constructed as:
\begin{equation*}
    g^{*} = \{x^{*}, y^{*}, \theta^{*}, w^{*}, h\}
\end{equation*}

\subsection{Implementation Details}\label{ID}
The loss function consists of a location loss, an orientation loss, and an opening width loss:
\begin{equation*}
\begin{aligned}
&\mathcal{L}(M_{q}, M_{\theta}, M_{w},\hat{M_{q}}, \hat{M_{\theta}}, \hat{M_{w}}) = \beta * \mathcal{L}_{loc}(M_{q}, \hat{M_{q}}) \\ &+ \gamma * \mathcal{L}_{ori}(M_{\theta}, \hat{M_{\theta}}) + \mathcal{L}_{width}(M_{w}, \hat{M_{w}})
\end{aligned}
\end{equation*}
where $\mathcal{L}_{-}$ denotes binary cross entropy loss, and $\hat{M_{q}}$, $\hat{M_{\theta}}$, and $\hat{M_{w}}$ are ground truth maps. The model is trained on a single NVIDIA RTX 3090 GPU for 500 epochs with a batch size of 1. We use Adam \cite{kingma2014adam} as the optimizer with an initial learning rate of $10^{-4}$ and weight decay. 

During training, two CLIP pre-trained encoders are frozen. At each iteration, we randomly sample an input-output tuple $d_j$ from $\mathcal{D}$.


\begin{figure}[t]
  \centering
  \begin{tikzpicture}[inner sep = 0pt, outer sep = 0pt]
    \node[anchor=south west] (fnC) at (0in,0in)
      {\includegraphics[height=1.12in,clip=true,trim=0in 0in 0in 0in]{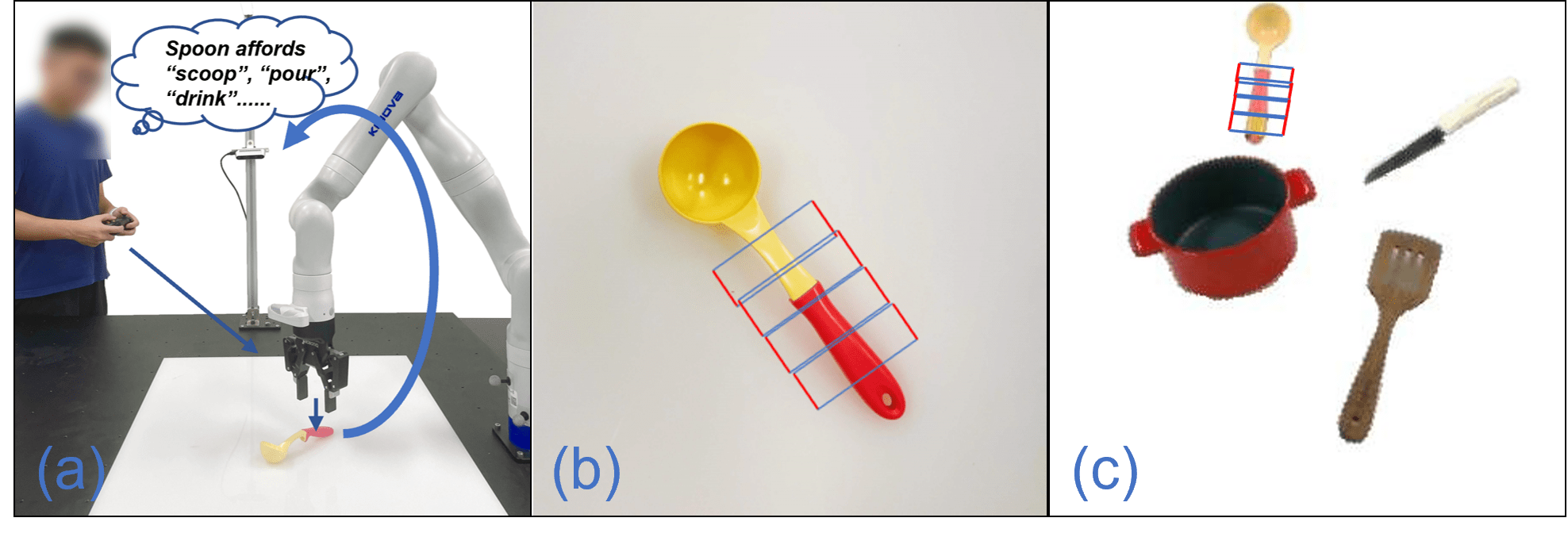}};
  \end{tikzpicture}
    \vspace*{-0.18in}
  \caption{Data generation: (a) A human
operator teleoperates the robot arm to stable grasp poses and assigns task labels. (b) 5D grasp poses on a single object are collected. (c) A multi-object scene with ground truth task-oriented grasp annotations are generated automatically.}
  \label{fig:data}
  \vspace*{-0.25in}
\end{figure}

\section{Dataset}\label{data}

To evaluate the performance of our design and established methods, a dataset $\mathcal{D}$ is required. Since there are no such datasets in the context of VLG, we build a custom one in two steps, including multi-object scene synthesis and template-based instruction generation.


In multi-object scene synthesis, we first crowdsource a list of object categories and tasks from four highly cited VG datasets: ContactDB\cite{brahmbhatt2019contactdb}, SG14000\cite{liu2020cage}, TOG-Net\cite{fang2020learning}, and TaskGrasp\cite{murali2020same}. Note that we are particularly interested in kitchen tools as they are frequently manipulated by an assistive robot. Full object set and task set can be found in our presentation video. Then, a human operator teleoperates a robot arm (see Fig.\ref{fig:data}(a)) to collect single-object grasping data (see Fig.\ref{fig:data}(b)). Each grasp may afford one or more tasks. Teleoperation allows for the extraction of tool grasping skills from real human behavior, without the significant risk of a sim-to-real gap that may arise when using simulated data. Additionally, an assistive robot usually perceives more than one object (i.e., target objects + distractors) in real-world applications. Therefore, similar to domain randomization \cite{tobin2017domain}, we randomly drop single-object data on synthetic backgrounds (see Fig.\ref{fig:data}(c)) to generate multi-object scenes with ground truth grasp annotations. This process is done automatically.

We apply a template-based instruction generation strategy to efficiently create $I_j$ at each iteration. 11 templates are adapted from \cite{nguyen2020robot}. Similar to \cite{chen2021joint}, we further augment the templates with QuillBot, an automatic paraphraser, to enrich the vocabulary and grammatical diversities. There are two types of instructions: (1) task with a target object (e.g., ``Use \textit{obj} to \textit{task}"), and (2) task only (e.g., ``Use something to \textit{task}"). Finally, \textit{obj} and \textit{task} are substituted with a target object category label and a target task label, respectively. Tab.\ref{tab:dataset} provides additional details of $D$.

\begin{table}[t]
\centering
\renewcommand\arraystretch{1.5}
\begin{tabular}{lc}
\hline
Parameters             & Settings                                  \\ \hline
Num of Categories      & 28                                       \\
Num of Instances       & 96                                       \\
Num of Tasks           & 38                                       \\
Num of Templates       & 106                                      \\
Grasp Type             & Planar                                   \\
Num of Grasps          & 10 Per Instance                          \\
Examples of Categories & Spoon, Fork, Mug, Pan, Scissor,  Tong    \\
Examples of Tasks      & Cut, Brush, Dig, Scoop, Handover, Saute  \\
Example of Templates   & ``Hold \textit{obj} in your hand and \textit{task}"                                  \\
Data Split Type        & Scene, Instance, Category, Category-Task \\ \hline
\end{tabular}
    \caption{Details of generated dataset}
    \label{tab:dataset}
        \vspace*{-0.4in}
\end{table}

\section{Experimental Setup} \label{exp_setup}

\subsection{Perception Experiments} \label{perception_exp_setup}

We evaluate the proposed method and baselines under four different test settings. The performance is measured by the ability to generalize to novel scenes, instances, object categories, and category-task combinations. For each level of generalization, the data is split into 80\% for training and 20\% for testing. Manually annotated grasps are used to evaluate models trained on $D$.

Four established baselines retrained on $\mathcal{D}$ are compared. The details are as follows:
\begin{itemize}
    \item \textbf{TAG} represents task-agnostic VG methods \cite{chu2018real, mousavian20196, mahler2017dex} that focus on grasp stability and ignores task suitability. It receives only visual inputs and randomly ranks each candidate's  task suitability. We remove the language component of GraspCLIP to model TAG.
    \item \textbf{CLIP+TAG} is a naive combination of CLIP and a task-agnostic grasp detector.
    It is originally introduced in \cite{gadre2022clip} for object localization. CLIP+TAG follows a two-stage pipeline where the task instruction is first grounded via Grad-CAM\cite{selvaraju2017grad} at the pixel level, followed by a standalone task-agnostic grasp detector.
    \item \textbf{OG+TAG} represents methods \cite{hatori2018interactively, shridhar2018interactive,zhang2021invigorate} focusing on grounding natural language to coarse object-centric representations. Specifically, we first ground the task instruction to the object bounding box with the highest matching score. A standalone task-agnostic grasp detector is then applied to that object.
    \item \textbf{CLIPort-S} is an adapted version of state-of-the-art visual-language manipulation and grasping framework CLIPort \cite{shridhar2022cliport}. We only keep its semantic branch and drop its spatial branch since depth information is unavailable. CLIPort does not explicitly consider task constraints when predicting grasps on the target object. 
\end{itemize}

\subsection{Real-Robot Experiments}
\begin{figure}[t]
  \centering
  \begin{tikzpicture}[inner sep = 0pt, outer sep = 0pt]
    \node[anchor=south west] (fnC) at (0in,0in)
      {\includegraphics[height=1.9in,clip=true,trim=0in 0in 0in 0in]{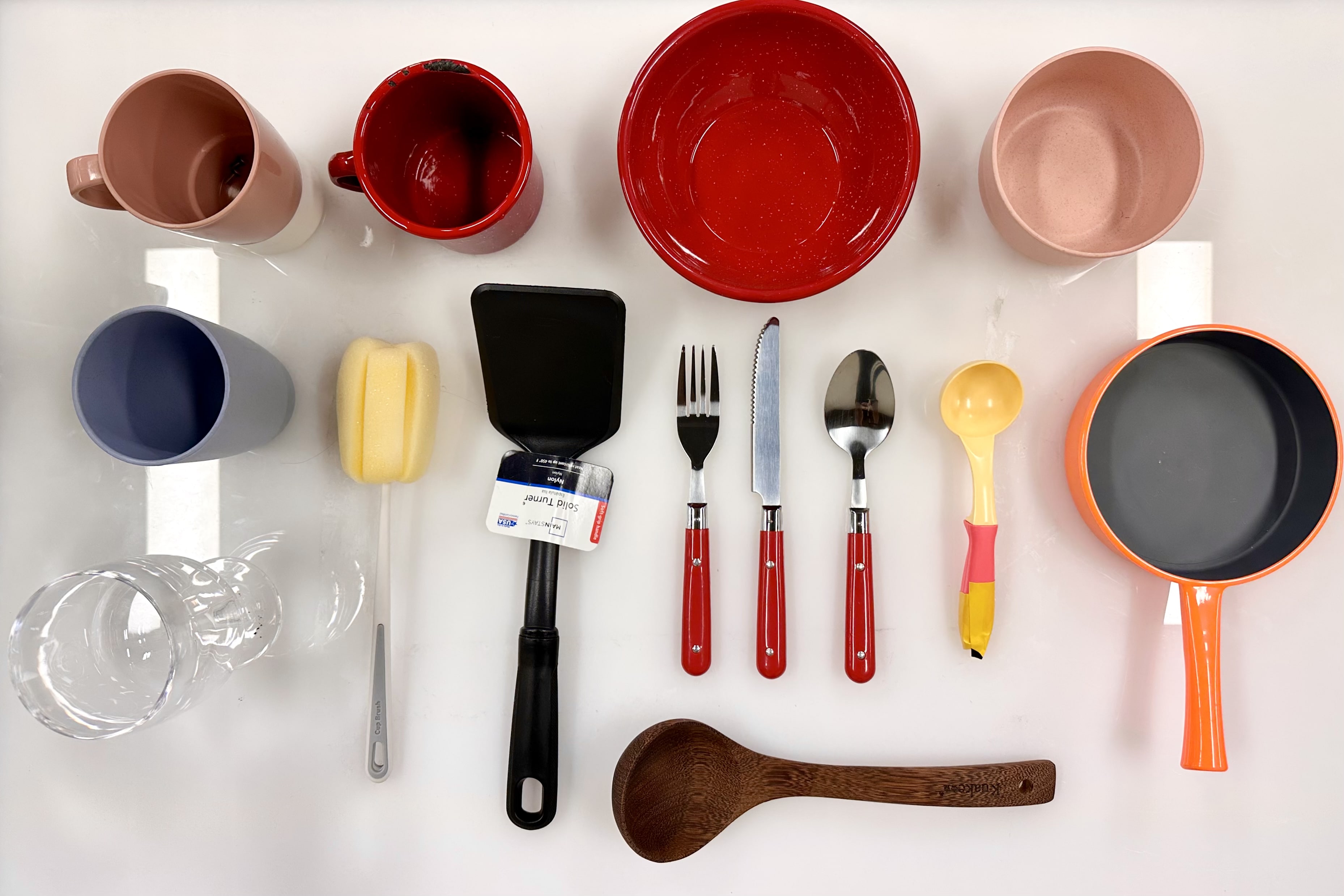}};
  \end{tikzpicture}
  \caption{Part of test objects collected from the laboratory and YCB dataset.}
  \label{fig:ycb}
  \vspace*{-0.25in}
\end{figure}

To further validate the effectiveness in real-world robotic applications, we deploy GraspCLIP on a 7-DoF Kinova Gen3 robot arm equipped with a Robotiq 2-finger adaptive gripper. Test objects are selected from the same categories as the training data but unseen during training. Some test kitchen tools collected from our laboratory and YCB dataset are shown in Fig.\ref{fig:ycb}. Here, we are only interested in revealing the gap between perception and execution. Therefore, the four baselines in Section \ref{perception_exp_setup} are not physically evaluated.


Converting the predicted grasp pose from image space to robot coordinate involves a sequence of transforms:
\begin{equation*}
    g_{robot} = T_{RC}(T_{CI}(g_{img}))
\end{equation*}
where $g_{robot}$ and $g_{img}$ are grasp poses in image space and robot coordinate, respectively. $T_{CI}$ transforms from 2D image space to 3D camera frame, and $T_{RC}$ transforms from camera frame to robot coordinate. 

The experimental procedure is as follows: (1) a set of $N$ objects ($N\geq1$) are placed in the robot workspace; (2) a natural language instruction is sent to the robot; and (3) the robot uses GraspCLIP to predict a task-oriented grasp pose $g$ and execute it on the target object.

\subsection{Evaluation Metrics}
Perception experiments examine the correctness of output grasps, while real-robot experiments test how well the robot physically interacts with objects. Two sets of evaluation metrics are adopted accordingly:
\begin{itemize}
    \item \textit{Perception experiments}: Following previous works \cite{mahler2017dex}\cite{chu2018real}\cite{wang2021high}, we consider a predicted grasp $g$ correct if two criteria are met: (1) the difference between the angle of $g$ and a ground truth grasp pose $ \hat{g}$ is less than $30^{\circ}$ and (2) the Jaccard index (similar to IOU) between $g$ and $ \hat{g}$ is greater than 0.25. The Jaccard index $J$ is defined as:
    \begin{equation*}
        J(g, \hat{g}) = \frac{| \hat{g} \cap g|}{| \hat{g} \cup g|}
    \end{equation*}
    Here, we choose the top-1 grasp candidate.
    \item \textit{Real-robot experiments}: To systematically evaluate the performance in real-robot experiments, we divide the pipeline into three stages and record their statistics separately. Three stages include Perception ($Perc$), Planning ($Plan$), and Action ($Act$). A grasp is considered successful if the target object is grasped subject to the task requirement and lifted stably for three seconds by the robot.
\end{itemize}

\section{Results} \label{exp}

\subsection{Result of Perception Experiments}
The result of perception experiments is reported in Tab.\ref{tab:sim_result}. Scene and instance-level generalizations focus on generalizing to cases with limited variances with respect to the training data. The \textbf{scene-level generalization} experiment creates novel scene layouts with seen categories, tasks, and category-task combinations. TAG randomly explores scenes without considering task instructions, setting a lower performance bound of 38.77\%. By incorporating CLIP, CLIP+TAG achieves a minor performance boost compared to TAG.  Although CLIP contains rich priors for grounding high-level concepts, it has a limited ability to inform low-level grasping directly. OG+TAG can accurately ground the task instruction to the bounding box of the target object but is unable to predict task-oriented grasps. CLIPort-S is a competitive baseline, achieving a relatively high success rate of 80.19\%. It still falls behind GraspCLIP since it does not explicitly consider the fine-grained effects of tasks on object grasping. GraspCLIP outperforms all baselines on scene-level generalization. In terms of \textbf{instance-level generalization}, all methods bear a performance drop due to intra-category variance. Still, GraspCLIP achieves the best performance at 85.73\%. 

\textbf{Category-level generalization} aims to transfer knowledge learned from familiar tool categories to novel ones. For example, having been taught that \textit{`cup"} has the function of ``\textit{pour}", the robot can recognize the novel category \textit{``bowl"} affords the same function. Task-only language instruction is used solely in this experiment. Any object that affords the task can be counted as the target object. Therefore, it increases the probability of TAG predicting correct grasps. OG+TAG suffers from detecting novel categories, performing poorly on this evaluation. GraspCLIP outperforms the second-best CLIPort-S by 4.43\%. \textbf{Category-task-level generalization} is a challenging though practical evaluation. A user may teach category-task pairs $spoon-scoop$ and $ladle-dispense$, and the robot should be able to mix the knowledge from two sources (i.e., $spoon-dispense$, $ladle-scoop$). GraspCLIP outperforms all the baselines. The performance on these two generalizations demonstrates the superiority of GraspCLIP in generalizing to relatively significant variances.

\begin{figure*}[t]
  \centering
  \vspace*{-0.2in}
  \begin{tikzpicture}[inner sep = 0pt, outer sep = 0pt]
    \node[anchor=south west] (fnC) at (0in,0in)
      {\includegraphics[height=2.8in,clip=true,trim=0in 0in 0in 0.0in]{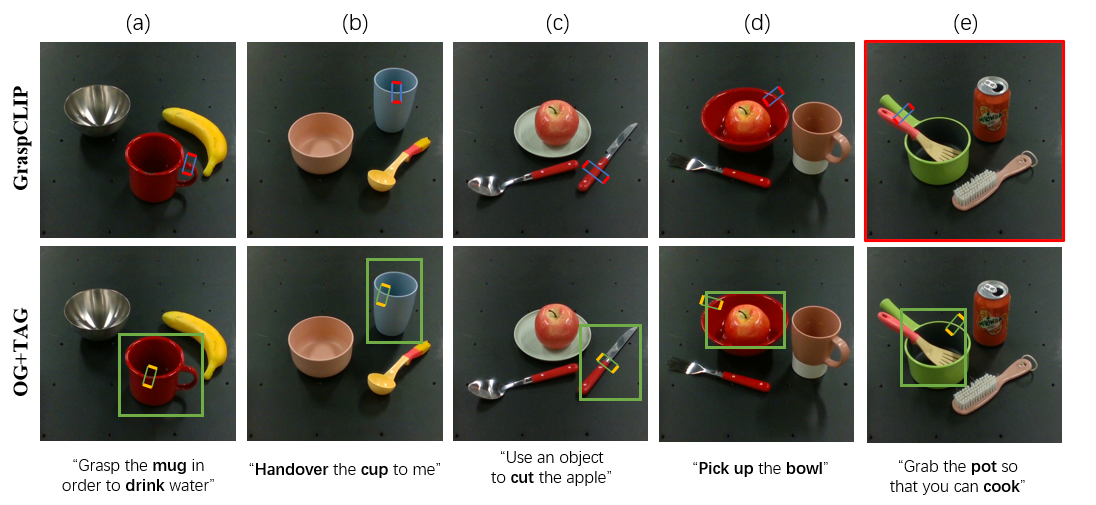}};
  \end{tikzpicture}
    \vspace*{-0.1in}
  \caption{Qualitative results of real-robot experiments. The grasps predicted by GraspCLIP and OG+TAG are represented by red-blue rectangle and yellow-green rectangle, respectively. The green boxes represent the bounding boxes detected by OG+TAG.}
  \label{fig:qualitative}
  \vspace*{-0.10in}
\end{figure*}

\subsection{Result of Real-Robot Experiments}
Real-robot experiments reveal the performance gap between perception and physical grasping. 
 Tab.\ref{tab:real_result} presents the quantitative results, and Fig.\ref{fig:qualitative} illustrates the qualitative results. Although baselines are not statistically evaluated in real-robot experiments, we provide the qualitative results of a representative baseline in Fig.\ref{fig:qualitative} for comparison. GraspCLIP achieves a high success rate in no clutter or lightly cluttered scenes (Fig.\ref{fig:qualitative}(a)-(c)). One of the limitations is low grasping DoF. While humans can perform 6 DoF grasping, such as grasping along $x$ or $y$ axis, GraspCLIP can only predict planar grasps (i.e., along $z$ axis). We plan to extend our framework to 6 DoF dexterous VLG.

We deliberately create complex scene layouts in the four-object setup to further gauge the limits of the implementation. The model performs reasonably well when the structure of the target object remains fair visibility. Some heavily cluttered layouts, such as stacking and containing (Fig.\ref{fig:qualitative}(d)-(e)), are hard to deal with. Therefore, the robot fails in some cases. Fig.\ref{fig:qualitative}(e) shows a failure case, highlighted in the red box on the rightmost side. A potential solution to this problem could involve equipping GraspCLIP with an active exploration module as in \cite{liu2020interactive}. Another type of failure comes from the language grounding error. In this case, the robot grasps a distractor with the same function as the target object. For example, when the task instruction is ``Use the \textit{laundry brush} to \textit{clean}", the robot falsely grounds the instruction to a sponge brush next to the laundry brush. Interactive correction with natural language \cite{sharma2022correcting
} could fix this error.

\begin{table}[t]
\renewcommand\arraystretch{1.4}
\centering
\begin{tabular}{ccccc}
\bottomrule
\multirow{2}{*}{Method} & \multicolumn{4}{c}{Generalization Type}    \\ \cline{2-5} 
                        & Scene & Instance & Category & Category-Task \\ \midrule
TAG          & 38.77 & 37.87    & 42.38  & 35.10   \\
CLIP+TAG   & 43.76 & 43.43    & 41.56   &  44.72       \\
OG+TAG   & 63.42 & 62.40    & 56.35  & 59.66       \\
CLIPort-S    & 80.19 & 75.01  & 78.80   & 75.99       \\
\midrule
GraspCLIP (Ours)   & \textbf{88.02} & \textbf{85.73}    & \textbf{83.23}  & \textbf{82.54}   \\ \bottomrule
\end{tabular}
\caption{Result of Perception Experiments (\%)}
\label{tab:sim_result}
  \vspace*{-0.3in}
\end{table}

\begin{table}[t]
\centering
\renewcommand\arraystretch{1.5}
\begin{tabular}{ccccc}
\bottomrule
\multirow{2}{*}{Num of Objects} & \multicolumn{4}{c}{Success} \\ \cline{2-5} 
                        & Perc   & Plan  & Act   & Overall  \\ \hline
$N=1$      & 19/20   & 19/20  & 19/20  & 95.00\%    \\
$N=2$         & 19/20   & 18/20  & 18/20  & 90.00\%    \\
$N=4$ (light clutter)         & 18/20   & 18/20  & 17/20  & 85.00\%    \\
$N=4$ (heavy clutter)       & 15/20   & 14/20  & 14/20  & 70.00\%    
\\ \bottomrule
\end{tabular}
\caption{Result of Real-Robot Experiments}
\label{tab:real_result}
  \vspace*{-0.4in}
\end{table}

\begin{table}[t]
\renewcommand\arraystretch{1.4}

\centering
\begin{tabular}{ccccc}
\bottomrule
\multirow{2}{*}{Method} & \multicolumn{4}{c}{Generalization Type}    \\ \cline{2-5} 
                        & Scene & Instance & Category & Category-Task \\ \midrule
COG-only            & 72.38 & 70.87   & 72.68  & 63.58 \\
FAG-only          & 77.48 & 74.07   & 77.93  & 73.14  \\
FAG-COG           & 68.31 & 64.39    & 58.37  &  61.93  \\
\midrule
RN50+BERT  & 82.68 & 78.49 & 82.80  & 80.74        \\
\midrule
GraspCLIP (Ours)  & \textbf{88.02} & \textbf{85.73}    & \textbf{83.23}  & \textbf{82.54}  \\ \bottomrule
\end{tabular}
\caption{Result of Ablation Studies (\%)}
\label{tab:as_result}
  \vspace*{-0.4in}
\end{table}

\subsection{Ablation Study}
To gain further insights into the effectiveness of each component, we conduct two sets of ablation studies. The result is reported in Tab.\ref{tab:as_result}.
\subsubsection{TOF Module Structure} 
\
\newline
\indent The proposed TOF module consists of a COG module and a FAG module. To investigate their effectiveness, we test three ablated versions shown in the first three rows of Tab.\ref{tab:as_result}.

COG-only uses two consecutive COG modules without task-level grounding, and FAG-only uses one FAG module without object grounding. Using two FAG modules gives a meaningless result, thus not reported. We observe that FAG-only performs consistently better than COG-only. This suggests that (1) task grounding is more critical for task-oriented grasp prediction, and (2) FAG is able to perform object localization to some extent. GraspCLIP outperforms both COG-only and FAG-only by a large margin. FAG-COG reverses the order of two grounding modules. The significant performance gap between FAG-COG and COG-FAG (i.e., GraspCLIP) justifies our coarse-to-fine design.

\subsubsection{CLIP-Based Encoders}
\
\newline
\indent To highlight the effectiveness of CLIP-based encoders over alternative pre-trained models, we substitute CLIP pre-trained encoders with an ImageNet-pretrained ResNet50 with BERT, denoted as RN50+BERT. In the fourth row of Tab.\ref{tab:as_result}, we observe that CLIP-based encoders consistently improve the performance across four generalization types, validating their effectiveness.




\section{Conclusion} \label{conclusion}

To address the challenge of task grounding in addition to object grounding in the context of VLG, GraspCLIP is proposed to enable task-oriented grasp prediction with visual-language inputs. Evaluation on a custom dataset demonstrates that GraspCLIP outperforms established baselines with object grounding only. To further validate the effectiveness, we deploy GraspCLIP on an assistive robotic arm for grasping previously unseen kitchen tools given the task specification. 
As a future direction, we consider the incorporation of interactive language correction into the GraspCLIP framework, as well as an extension of GraspCLIP to support 6 DoF dexterous VLG.












\bibliographystyle{IEEEtran}
\balance
\bibliography{main}

\end{document}